\documentclass[letterpaper, 10 pt, conference, nofonttune]{ieeeconf} 

\IEEEoverridecommandlockouts                              

\usepackage{amsthm}

\usepackage{subfigure}
\usepackage{xcolor}
\usepackage{times}
\usepackage{epsfig}
\usepackage{amsmath}
\usepackage{amssymb}
\usepackage{mathrsfs}
\usepackage[ruled,vlined,linesnumbered]{algorithm2e}
\usepackage{wrapfig}
\usepackage{hyperref}

\newcommand{\GGG}{\mathcal{G}}

\newcommand{\EEE}{\mathcal{E}}

\newcommand{\VVV}{\mathcal{V}}

\newcommand{\OOO}{\mathcal{O}}

\newcommand{\MMM}{\mathcal{M}}

\newcommand{\hh}{\mathbf{h}}

\newcommand{\vv}{\mathbf{v}}
\newcommand{\qq}{\mathbf{q}}

\newcommand{\pp}{\mathbf{p}}

\newcommand{\MM}{\mathbf{M}}
\newcommand{\bb}{\mathbf{b}}

\newcommand{\WW}{\mathbf{W}}

\makeatother

\title{Collaborative Decision-Making Using Spatiotemporal Graphs in \\ Connected Autonomy
}

\newcommand{\RRR}{\mathbb{R}}
\newcommand{\RR}{\mathcal{R}}

\begin{document}
\author{Peng Gao, Yu Shen, and Ming C. Lin
\thanks{Peng Gao, Yu Shen, and Lin C. Lin are with the Department of Computer Science, University of Maryland, College Park, MD 20742, USA. {Email: \{gaopeng, yushen, lin\}@umd.edu}.}%
}
\maketitle


\begin{abstract}
Collaborative decision-making is an essential capability for multi-robot systems, such as connected vehicles, to collaboratively control autonomous vehicles in accident-prone scenarios. Under limited communication bandwidth, capturing comprehensive situational awareness by integrating connected agents' observation is very challenging. In this paper,  we propose a novel collaborative decision-making method that efficiently and effectively integrates collaborators' representations to control the ego vehicle in accident-prone scenarios. Our approach formulates collaborative decision-making as a classification problem. We first represent sequences of raw observations as spatiotemporal graphs, which significantly reduce the package size to share among connected vehicles. Then we design a novel spatiotemporal graph neural network based on heterogeneous graph learning, which analyzes spatial and temporal connections of objects in a unified way for collaborative decision-making. We evaluate our approach using a high-fidelity simulator that considers realistic traffic, communication bandwidth, and vehicle sensing among connected autonomous vehicles. The experimental results show that our representation achieves over $100$x reduction in the shared data size that meets the requirements of communication bandwidth for connected autonomous driving. In addition, our approach achieves over  $30\%$ improvements in driving safety. 
\end{abstract}


\section{Introduction}

Multi-robot systems have received considerable attention over the past few decades, due to their remarkable attributes of redundancy \cite{gao2023collaborative}, scalability \cite{gao2023uncertainty}, and parallelism \cite{reily2021adaptation}. Among these, the connected autonomous vehicle stands out as a prominent example of collaborative multi-robot systems. Unlike conventional studies that focus on single-robot scenarios, connected autonomous driving considers the collective capabilities of multiple autonomous vehicles, enabling enhanced performance and efficiency in diverse tasks, such as object detection \cite{zhao2021point}, tracking \cite{li2021learning} and autonomous driving control \cite{cui2022coopernaut}

To enable efficient connected autonomous driving, collaborative decision-making is a fundamental ability, with the goal of enabling connected vehicles to efficiently share and utilize observations provided by ego and collaborator vehicles, thus mitigating blind spots and collaboratively making optimal decisions (e.g., taking brake actions), especially in accident-prone scenarios. As shown in Figure, the ego vehicle (shown in gray) is blocked by yellow cars, and it may be difficult for the car's sensors to detect an approaching red car rushing through the intersection. In this case, by integrating observations provided by collaborative vehicles, the ego vehicle eliminates its own blind spots and takes a "brake" action to avoid a traffic accident. 
However, collaborative decision-making in connected autonomous driving is very challenging, as the communication bandwidth between vehicles is limited, which does not allow connected vehicles to directly share their raw observations \cite{kenney2011dedicated, gallo2013short}. In addition, occlusion in observations, highly dynamic street conditions, and the complex interactions of objects make collaborative decision-making very hard to solve.

\begin{figure}[t]
\vspace{6pt}
\centering
\includegraphics[width=0.455\textwidth]{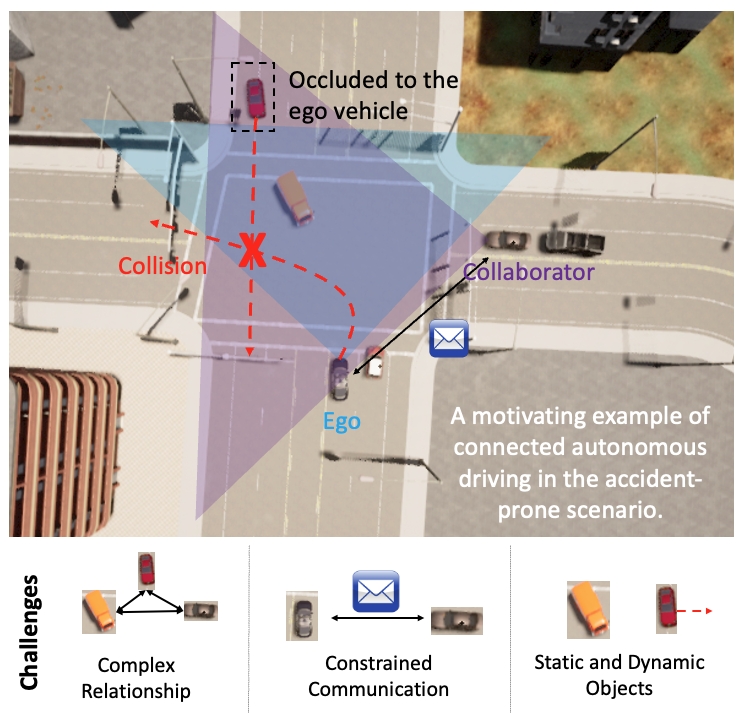}
\vspace{-8pt}
\caption{An motivating example of connected autonomous driving in the accident-prone scenario. To avoid accidents, connected vehicles need to effectively share and integrate their own observations, meanwhile addressing the challenges caused by communication constraint and complex interactions among static and dynamic street objects.
}
\vspace{-12pt}
\label{fig:motivation}
\end{figure}

Various methods are studied to address the challenges, which can be generally divided into three groups, including raw-based early collaboration, output-based late collaboration, and feature-based intermediate collaboration methods. 
Raw-based early collaboration utilizes multi-robot raw observations, which generates the most comprehensive situational awareness. These methods can effectively remove blind spots of individuals, but far beyond communication bandwidth limitation in outdoor applications \cite{han2023collaborative}.
Output-based late collaboration methods focus on sharing and fusing multi-robot prediction results \cite{forsyth2014object,song2023cooperative}, which significantly reduces the consumption of communication bandwidth. However, these results are predicted based on single-robot observations with incomplete information about a scene, which are highly to be inaccurate and noisy.
Feature-based intermediate collaboration methods mainly focus on utilizing the extracted features for collaboration tasks, such as sharing downsampling lidar points \cite{zhao2021point} or extracting features via PointNet \cite{cui2022coopernaut}. These approaches achieve a trade-off between communication efficiency and performance. However, how to compactly preserve the cues of a scene for sharing and aggregation, such as temporal cues, has not been well addressed yet.


In this paper, we propose a novel collaborative decision-making method that efficiently integrates spatiotemporal observations provided by connected vehicles for autonomous driving. First, we represent each vehicle's observation sequence as a {\em spatiotemporal graph}, with the nodes to encode the locations of the detected objects, the spatial edges to encode the spatial distance between pairs of objects, and temporal edges to encode the motion of objects. Given the spatiotemporal representations, connected vehicles can efficiently share sequential observations while overcoming communication constraints. Then, we merge all ego-collaborator representations given the Global Navigation Satellite System (GNSS) poses and formulate collaborative decision-making as a classification problem.  We utilize {\em heterogeneous graph learning} as a framework that simultaneously analyzes spatial and temporal relationships of objects, thus encoding comprehensive situational awareness. Given the situational embedding, we predict if the ego vehicle should take brake action to avoid an accident or not. The full approach is learned by imitation learning with expert actions.

Our key contribution is the introduction of a novel collaborative decision-making method for connected autonomous driving. Specifically,
\begin{itemize}
    \item We propose a novel representation based on {\em spatiotemporal graphs} generated from a sequence of observations, which integrate not only the current states but also the historical motion of street objects. Our representation achieves a greater than {\bf 100x} reduction in the shared data size that meets the requirements of communication bandwidth for connected autonomous driving.

    \item We present a novel {\em spatiotemporal graph neural network} based on {\em heterogeneous graph learning}, which generates the embedding of the spatiotemporal graph by analyzing the spatial and temporal connection of objects in a unified way, thus encoding cues for collaborative decision-making. Our approach achieves over {\bf 30$\%$} improvements in driving safety.

\end{itemize}

\section{Related Work}
Connected autonomous driving based on collaborative perception provided by connected agents has attracted extensive attention recently. The existing methods can be generally divided into three groups, including raw-based early collaboration, output-based late collaboration, and feature-based intermediate collaboration methods.

The early collaboration fuses the raw data for the input of the network, which requires connected agents to share, transform, and aggregate raw sensor data onboard for vision task  \cite{arnold2020cooperative, chen2019cooper}. 
The late collaboration usually adopts fusion at the postprocessing stage, which merges multi-agent perception outputs, such as Non-Maximum suppression to remove redundant prediction \cite{forsyth2014object} and refined matching to remove results that violate the pose consistency \cite{song2023cooperative}. 
The intermediate collaboration aims to learn and share compressed features from the raw observations, which is a trade-off between communication bandwidth and performance. From the data-sharing perspective, different communication mechanisms are developed, such as When2com \cite{liu2020when2com}, Who2com \cite{liu2020who2com}, and Where2com \cite{hu2022where2comm}. From the data fusion perspective, the strategies include direct concatenation \cite{chen2019f}, re-weighted sum \cite{guo2021coff}, graph learning-based fusion \cite{wang2020v2vnet, zhou2022multi}, and attention-based fusion \cite{xu2022v2x,xu2022opv2v}. 
From the task perspective, various tasks are studied, including object detection \cite{bi2022edge}, tracking \cite{li2021learning}, semantic segmentation \cite{xu2022cobevt}, localization \cite{yuan2022keypoints}, depth estimation \cite{hu2023collaboration} and autonomous driving control \cite{cui2022coopernaut}.

The early collaboration contains the most comprehensive information of a scene, which can overcome occlusion and long-distance observations, but these methods can not be used in outdoor environments due to their large bandwidth requirements. Late collaboration significantly reduces the communication bandwidth cost by directly sharing outputs, however, as individual observations are often noisy or incomplete, late collaboration generally has the worst performance. Intermediate collaboration balances communication efficiency and collaboration performance. However, these methods do not have redundant bandwidth to share a sequence of features, which causes the ignoring of temporal cues for collaborative decision-making.

To encode spatial and temporal information for decision-making, spatiotemporal graph learning is widely used in single-robot scenarios, such as trajectory prediction \cite{yu2020spatio}, path planning \cite{meliou2007nonmyopic}, object localization \cite{gao2022asynchronous, gao2021multi}, and reasoning \cite{liu2020spatiotemporal}. These methods generally treat the embedding of the spatial and temporal domain separately, such as using a recurrent neural network to aggregate temporal information and using a graph neural network to aggregate spatial information \cite{seo2018structured, seo2018structured,chen2022gc}, or alternating temporal and spatial blocks during convolution \cite{yu2017spatio, guo2019attention, zheng2020gman}. 
Even though these approaches have achieved promising results in vision tasks, they typically assume that a person or vehicle can be continuously tracked, and use a fixed-length observation sequence as input. Therefore these methods are difficult to apply to collaborative decision-making with the observed objects being prone to intermittent losses within the observation sequence and the available sequence length being variant. In this paper, we propose a novel spatiotemporal graph network based on heterogeneous graph learning, which can deal with length-variant observations with objects missing, meanwhile preserving spatial and temporal connections of objects for connected autonomous driving.

\begin{figure*}[t]
\centering
\includegraphics[width=0.975\textwidth]{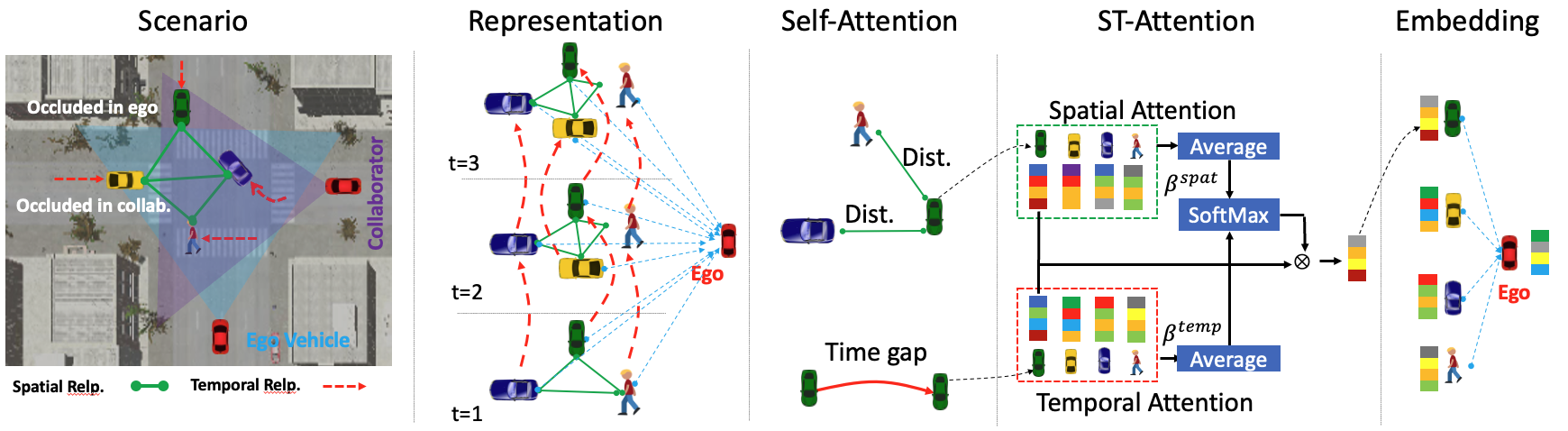}
\caption{An overview of our approach of spatiotemporal graph embedding for connected autonomous driving. An ego vehicle and a collaborator vehicle meet at an intersection, each of them has occluded objects in their own field of view. We represent their observation sequences as spatiotemporal graphs and merge them all to generate the final representation of the scene. Our spatiotemporal graph learning network captures spatial and temporal relationships of objects by analyzing the importance of spatial and temporal attention and, finally, generating the situational embedding of the scene with respect to the ego vehicle.
}
\label{fig:approach}
\vspace{-6pt}
\end{figure*}
\section{Approach}

\textbf{Notation.} Matrices are represented as boldface capital letters, e.g., $\MM = \{\MM_{i,j}\}\in \RR^{n\times m}$. $\MM_{i,j}$
denotes the element in the $i$-th row and $j$-th column of $\MM$.
Vectors are denoted as boldface lowercase letters $\vv \in \RR^{n}$ and scalars are denoted as lowercase letters.

\subsection{Problem Formulation}
We propose a collaborative decision-making approach for connected autonomous driving in accident-prone scenarios. Formally, we assume that there is one ego vehicle and $n-1$ collaborative vehicles that receive RGB-D or Lidar observations.
Each vehicle provides a sequence of observations $\OOO_k = \{obs_k^t, obs_k^{t+1}, \dots, obs_k^{t+T} \}, k \in \{0, 1,2, \dots, n\}$.
Each observation recorded at time $t$ consists of detected objects $obs^t = \{\vv_1^t, \vv_2^t, \dots, \vv_m^t\}$ where $\vv_i \in \RRR^3$ denotes the 3D position of the $i$-th object detected at time $t$. 

Given a sequence of observations $\mathcal{O}_k$, 
we represent it as a spatiotemporal graph $\GGG_k=\{\VVV_k,\EEE_k^{spat},\EEE_k^{temp}\}, k=1,2,\dots,n$. $
\VVV_k = Unique(\OOO_k)$ denotes the node set, which contains all the positions of objects detected by $n$ vehicles, where $Unique$ denotes the unique operation that removing the duplicated objects in $\OOO_k$. In addition,
$\EEE^{spat}_k = \{e_{p,q}^{spat}\}$ denotes the spatial relationships between a pair of 
objects.
$e^{spat}_{p,q}=||\vv^t_p-\vv^t_q||_2$ denotes the distance between the $p$-th object and the $q$-th object recorded at the same time $t$, otherwise $e^{spat}_{p,q} = 0$.
$\EEE^{temp}_k = \{e_{p,q}^{temp}\}$ denotes the temporal relationships of the same 
object recorded at different times. If $\vv_p^{t_1}$ and $\vv_q^{t_2}$ are the same object recorded at time $t_1$ and $t_2$, then
$e^{temp}_{p,q}=t_2-t_1$, otherwise $e^{temp}_{p,q}=0$.

Given the $n$ spatiotemporal graph representations provided by $n$ connected vehicles, we first transform the collaborator vehicles' observations to the ego vehicle's coordinate and merge them together with the ego observations, thus eliminating the ego vehicle's blind spots. The transformation of each collaborative vehicle is obtained from the GNSS sensor \cite{cui2022coopernaut}. The merged spatiotemporal graph is denoted as $\MMM = \phi(\GGG_{ego}, \GGG_1, \GGG_2, \dots, \GGG_m)$, where $\phi$ denotes the transformation and merge function and $m$ denotes the number of collaborator vehicles that are close to the ego vehicle within a distance threshold. The threshold is set to $150$ meters. As the ego vehicle may be observed by other vehicles, we add the ego node $obs_{ego} = [0,0,0]$ and remove all nodes within $2$ meters away from the ego node. Then, we fully connect the ego node to all the other nodes for message-passing purposes. Finally, the merged spatiotemporal graph is defined as $\MMM = \{\VVV,\EEE^{spat},\EEE^{temp}\}$, as shown in Figure \ref{fig:approach}.

Based upon the merged spatiotemporal graph $\MMM$, we formulate collaborative decision-making for connected autonomous driving as a classification problem. The goal is to identify if the current situation is dangerous or not, thus allowing the ego vehicle to take brake or driving actions. Given the spatiotemporal graph representations, we significantly reduce the amount of shared data packages between connected vehicles and preserve the important cues, including spatial and temporal relationships of street objects, for decision-making in accident-prone scenarios.


\subsection{Spatiotemporal Graph Embedding}
To obtain the embedding of spatiotemporal graphs for decision-making, we propose a heterogeneous graph attention network that encodes object locations and their spatiotemporal relationships in a unified way, as shown in Figure \ref{fig:approach}. 
The embedding of the spatiotemporal graph is defined as $\hh_{ego}^\prime = \psi(\MMM)$, where $\hh_{ego}^\prime$ is the embedding of the holistic scene with respect to the ego vehicle and $\psi$ is the heterogeneous attention network. Formally, we first project each node feature to the same feature space, which is defined as follows:
\begin{equation}\label{eq:project}
    \hh_i = \WW_v \vv_i 
\end{equation}
where $\hh_i$ denote the projected feature of the $i$-th node, $\WW_v$ denote the associating weight matrix. $\vv_i$ denotes the position of the $i$-th object. 
Then we compute the self-attention of each node given the spatial edges, defined as follows:
\begin{equation}\label{eq:self-att}
    \alpha_{i,j}^{spat} =  \frac{ \exp\left(\sigma ([\WW\hh_i||\WW\hh_j||\WW_e e_{i,j}])\right)}{\sum_{e_{i,k} \in \EEE^{spat}}\exp(\sigma ([\WW\hh_i||\WW \hh_k||\WW_e e_{i,k}]))} 
\end{equation}
where $\alpha_{i,j}^{spat}$ is the attention from node $j$ to node $i$, $\sigma$ denotes the ReLu activation function,  $||$ denotes the concatenation operation, $\WW$ and $\WW_e$ are weight matrices. 
This attention weight is obtained by comparing the query of the $i$-th node with
its neighborhood nodes meanwhile considering their edge attributes. The final attention is normalized by the SoftMax function.
To encode temporal relationships of objects, we can easily traverse all temporal edges in $\EEE^{temp}$ in Eq. (\ref{eq:self-att}). 
Finally, given different types of edges, we get two attentions $\alpha_{i,j}^{spat}, \alpha_{i,j}^{temp}$ for the $i$-th node. Then, we compute the node embedding vector as follows:

\begin{equation}\label{eq:group}
\hh_i^{spat} = \sigma \left(\WW\hh_i + \sum_{e_{i,k} \in \EEE^{spat}} \alpha_{i,j}^{spat} (\WW\hh_j +\WW_e e_{i,j}) \right)
\end{equation}
Based upon Eq. (\ref{eq:group}), we compute the $i$-th node  embedding vectors $\hh_i^{spat}$ and $\hh_i^{temp}$ given their associating attentions $\alpha_{i,j}^{spat}, \alpha_{i,j}^{temp}$. They are computed via aggregating the object embedding feature and their spatiotemporal edge attributes weighted by attention weights.
We also use a multi-head mechanism to enable the network to catch a richer representation of the embedding. Multi-head embedding vectors are concatenated after intermediate attention layers.

Given the spatiotemporal embedding of nodes, we further learn the importance of different types of relationships of objects. The importance of spatial relationships is computed as follows:
\begin{equation}\label{eq:beta_spat}
    \beta^{spat} = \frac{1}{|\VVV^{spat}|}\sum_{i \in \VVV^{spat}} \qq^T tanh(\WW_b\hh_i^{spat} +\bb)
\end{equation}
where $\WW_b$ denotes the weight matrix, $\bb$ is the bias vector, $\qq$ denotes the learnable edge-specific attention vector and $\VVV^{spat}$ denotes the node set containing nodes with spatial edge connections. The importance of spatial relationships is obtained by averaging the spatial embedding vectors of all nodes.
Similarly, the importance of temporal relationships is computed through Eq. (\ref{eq:beta_spat}) with the node set $\VVV^{temp}$, where $\VVV^{temp}$ denotes the node set containing nodes with temporal edge connections. The learnable parameters are shared for the computation of the importance of spatial and temporal relationships. Then, $\beta^{spat}$ and $\beta^{temp}$  are normalized through SoftMax, defined as follows:
\begin{equation}
    \beta^{spat} = \frac{\exp(\beta^{spat})}{\exp(\beta^{spat} + \beta^{temp})}
\end{equation}
\begin{equation}
    \beta^{temp} = \frac{\exp(\beta^{temp})}{\exp(\beta^{spat} + \beta^{temp})}
\end{equation}
where $\beta^{spat}, \beta^{temp}$ denotes the contribution of the type of relationship for decision-making. The higher the value, the larger the importance of the type of relationship. The final situational embedding vector with respect to the ego vehicle is computed as:
\begin{equation}\label{eq:final}
\hh_{ego}^\prime = \sum_{\Psi \in \{spat, temp\}} \beta^\Psi \hh^\Psi_{ego}    
\end{equation}
where $\hh_{ego}^\prime$ denotes the node embedding vector of the ego node, which integrates all object positions and their spatiotemporal relationships.


\subsection{Connected Autonomous Driving}
Given the status embedding of the ego vehicle $\hh_{ego}^\prime$, we predict the action of the ego vehicle to identify if the ego vehicle should take brake action or not. Formally, the prediction is defined as follows:
\begin{equation}\label{eq:mlp}
    \pp = \text{SoftMax}\left(\sigma\left(MLP([\hh_{ego}^\prime || cmd])\right)\right)
\end{equation}
where $ \pp =[p_1,p2]$ denotes the prediction with $p_1$ denoting the probability of taking brake and $p_2$ denoting the probability of keeping running, and $p_1+p_2 =1$.
$cmd$ denotes the command given to the ego vehicle, including lane follows, turn right, turn left, go straight, change left, and change right, which is represented as a one-hot vector. $\sigma$ denotes the $ReLu$ activation function and  $MLP$ denotes a multi-layer perceptron which contains one linear layer. Finally, the prediction result is normalized by the SoftMax function. To train our network, we use the cross-entropy loss.

\section{Experiment}

\begin{figure}[t]
\vspace{4pt}
\centering
\includegraphics[width=0.485\textwidth]{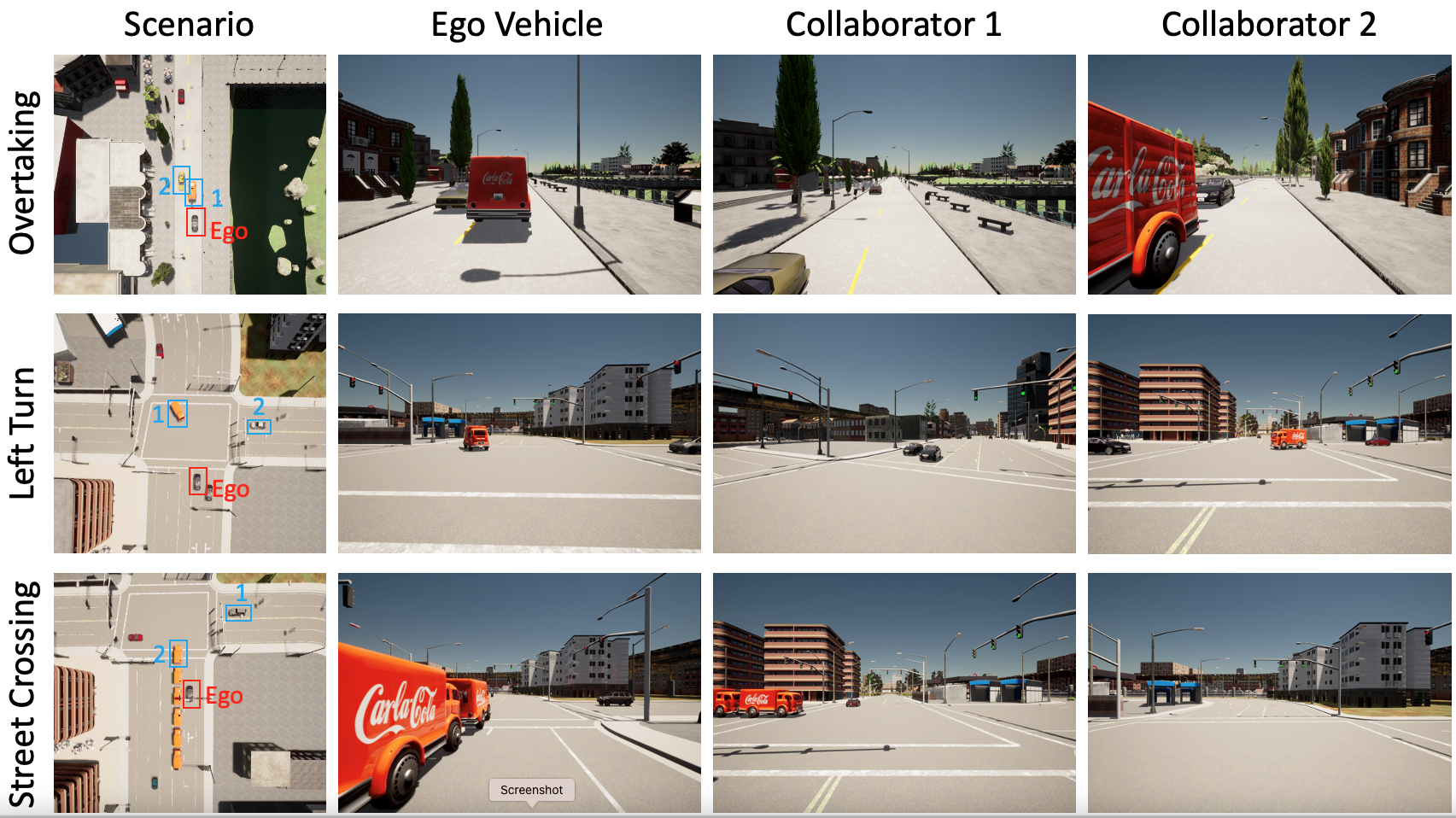}
\caption{Three scenarios of CAD, including overtaking, left turn, and street crossing. Each scenario contains one ego vehicle and at least two collaborator vehicles.  
}
\label{fig:scenario}
\end{figure}

\begin{table}[t]
\centering
\tabcolsep=0.2cm
\caption{Communication standard used in the simulation with the sensing frequency of $10$ Hz}
\label{tab:commu}
\begin{tabular}{|l|c|c|c|c|c|c|c|c|c|c|c|c|}
\hline
Method  & Bandwidth & Package Size  & Packet Loss \\
	\hline
DSRC \cite{kenney2011dedicated}
        & 2 Mbps & 200KB & $< 5\%$\\
\hline 
C-V2X \cite{gallo2013short} 
        & 7.2 Mbps & 720 KB & $<5\%$ \\
\hline 
\end{tabular}\vspace{-8pt}
\end{table}

\begin{table*}[t]
\vspace{8pt}
\centering
\tabcolsep=0.4cm
\caption{Quantitative Comparison of Different Methods for connected autonomous driving: \\  Accident Detection (\textbf{AD}) for driving safety and Expert action rate (\textbf{EAR}) for the imitation performance.  }
\label{tab:QuanResults}
\begin{tabular}{|l|c|c|c|c|c|c|c|c|c|c|c|c|}
\hline
Method  & \multicolumn{1}{ c|}{Package Size }  & \multicolumn{2}{ c|}{Overtaking }  &  \multicolumn{2}{ c|}{Left Turn }  & \multicolumn{2}{ c|}{Street Crossing}\\
\cline{2-8}
		& PS $\downarrow$ & AD $\uparrow$ & EAR $\uparrow$ 
         & AD $\uparrow$ & EAR $\uparrow$ 
         & AD $\uparrow$ & EAR $\uparrow$ \\
		\hline
        A) Raw Data Sharing \cite{zhao2021point}
        & 6 MB  & 0.7619 & 0.7126 & 0.3522 & 0.6440 & 0.3165 & 0.8650     \\
        \hline

        B) GAT \cite{velickovic2017graph}
        & 0.6KB &0.7588 & 0.6565 & 0.2272 & 0.6041 & 0.2960 & 0.7505\\
        \hline

        C) Collab-GAT \cite{xie2020mgat}
        &0.6 KB  & 0.5983 &0. 7020 & 0.2561 & 0.6204  & 0.5018 & 0.7195\\	
		\hline
        
		D) Compressed Feature Sharing \cite{cui2022coopernaut}
        & 510 KB  & 0.7066 & 0.7485 & 0.4962 & 0.6603  & {0.4690} & \textbf{0.8651}    \\
        \hline 

        \hline
		E) Ours w.o. Edge Attribute  & 4.8 KB
         & 0.7434  & 0.7322  & 0.4148 & 0.7427& 0.2564 & 0.7785\\
	\hline 
        F) Ours (All Elements)
        &4.9 KB & \textbf{0.9265}  & \textbf{0.8336} & \textbf{0.6070} & \textbf{0.7670}  & \textbf{0.6451} & {0.7846}\\
        \hline
        Improvements (F/D) 
        & 104.08 & 1.3224 & 1.1137 & 1.2233 & 1.1616 & 1.3755 & 0.9069 \\
        \hline
  
\end{tabular}
\vspace{-6pt}
\end{table*}

\subsection{Experimental Setups}
We employ a high-fidelity connected autonomous driving (CAD) simulator to evaluate our approach, which integrates CARLA \cite{Dosovitskiy17} and AutoCast \cite{autocast}.
CARLA is an open-source autonomous driving simulator that is able to simulate vehicle sensors and traffic scenarios. AutoCast is an end-to-end framework built upon CARLA, which provides V2V communication and sensor sharing, thus achieving collaborative decision-making and vehicle collaboration.
In our experiments, following the recent work \cite{cui2022coopernaut}, we utilize three different traffic scenarios at street interactions where accidents more frequently occur, as depicted in Figure \ref{fig:scenario}. Specifically,
\begin{itemize}
    \item \textbf{Overtaking}: A truck is obstructing a sedan's path on a two-way, single-lane road marked with a dashed yellow divider. Moreover, the truck is hindering the sedan's visibility of the oncoming lane. The autonomous vehicle (ego car) needs to perform a lane change maneuver to pass the truck. 
    \item \textbf{Left Turn}: An ego car attempts to make a left turn at a yield light. However, it faces an obstacle in the form of another truck positioned in the opposing left-turn lane. This obstacle limits the ego car's visibility of the lanes across from it, including any vehicles that might be proceeding straight.
    \item \textbf{Street Crossing}: While the ego car is crossing the street, another vehicle runs the red light. The sensing system is unable to detect this vehicle due to the presence of lined-up vehicles waiting to make a left turn.
\end{itemize}

For each scenario, we collect 24 trials, of which 12 trials are used for training and 12 trials are used for testing. Each trial contains 300 data instances and each data instance includes RGBD images observed by an arbitrary number of connected vehicles, the GNSS positions and orientations of vehicles, the command of ego vehicle, and the ground truth of vehicle actions. The data collection frequency is $10$ Hz.

We use a sequence of 15 frames for each spatiotemporal graph generation. we use YOLOv5 \cite{glenn_jocher_2022_6222936} to detect objects and use SORT \cite{bewley2016simple} to track objects. 
We extract the position as each node's attribute, which is obtained from the depth images.  The spatial edges of non-ego objects are fully connected and spatial edge attributes (distances) are calculated from pairs of objects' positions. The temporal edges are connected via the object tracking results and temporal edge attributes (time gap) are identified via the temporal tracking and timestamps.
The GNSS  positions and orientations of each vehicle are represented as a transformation matrix with respect to the world coordinate.

In the implementation of our network, the heterogeneous attentional graph network $\phi$ is implemented based on the PyTorch geometric library. 
We set the number of layers of the network to be $2$ and set $\WW_v \in \RRR^{4\times 12}$, $\WW \in \RRR^{12\times 6}$ $\WW_e \in \RRR^{1\times 6}$ and $\WW_b \in \RRR^{6\times 6}$ separately. In addition, we set $\qq \in \RRR^{6 \times 1}$ defined in Eq. (\ref{eq:group}) and the multi-head number is $4$. The MLP defined in Eq. (\ref{eq:mlp}) consists of three linear layers, the first two layers. In all the experiments, we use ADAM \cite{zhang2018improved} as the optimization method. We run $100$ epochs to train our approach.

For the comparison, we first implement a baseline method that is \textbf{our full approach but without considering edge attributes}, including object distances in spatial edges and time gaps in temporal edges. In addition, we compare our approach with four existing methods, including
\begin{itemize}
    \item \textbf{Raw Data Sharing} that directly integrates raw Lidar observations and utilizes the Point Transformer to extract features for collaborative decision-making \cite{zhao2021point}. As direct processing raw data on standard computers is unfeasible, we downsample the raw Lidar data size to 4096.
    \item \textbf{GAT} that uses graph representation generated from single-vehicle observations and graph attention neural network to encode the representation for connected autonomous driving \cite{velickovic2017graph}.
    \item \textbf{Collab-GAT} that is similar to GAT but uses multi-vehicle observations as input for collaborative decision-making in connected autonomous driving \cite{xie2020mgat}.
    \item \textbf{Compressed Feature Sharing} that extracts compressed features from multi-vehicle observations and integrates the compressed features via voxel pooling to encode the situational awareness for collaborative decision-making. 
\end{itemize}
None of these comparison methods can utilize temporal information due to the communication bandwidth constraint and the model design.

We employ three metrics for the evaluation of connected autonomous driving, including
\begin{itemize}
    \item Package Size (\textbf{PS}) is the size of the shared package between connected vehicles, which is used to evaluate communication efficiency.
    \item Accident Detection (\textbf{AD}) is defined as the ratio of detected accident-prone cases overall ground truth accident-prone cases. Given the ground truth control of the ego vehicle, the accident-prone case is when the ego vehicle takes brake actions.
    \item Expert Action Rate (\textbf{EAR}) that is defined as the ratio of correct reproduced expert actions over all the number of expert actions, which is used to evaluate the imitation performance.
\end{itemize}

\begin{figure}[t]
\vspace{6pt}
\centering
\includegraphics[width=0.485\textwidth]{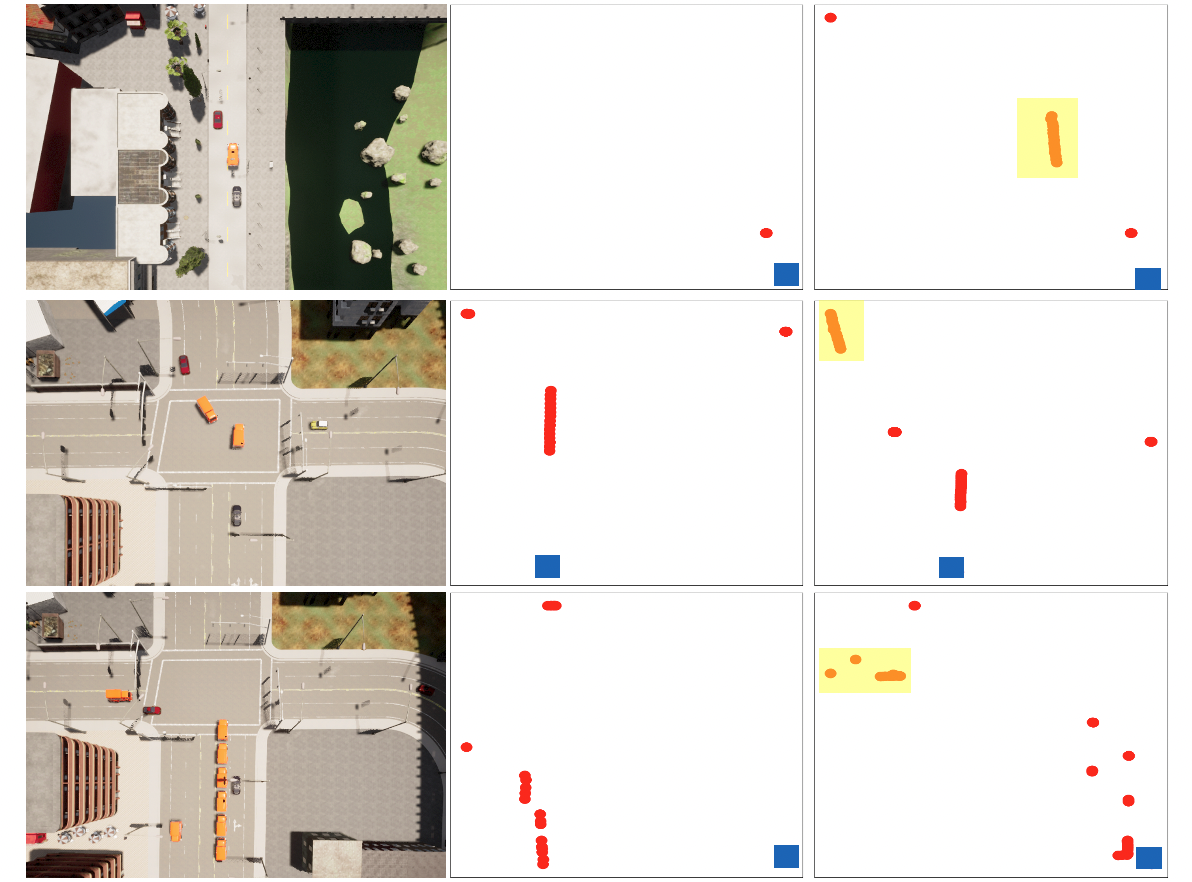}
\caption{Qualitative Results. The first column is the detected accident by our approach. The second column is the representations (ignore edges for simplification) of ego observation sequences, where the blue rectangle denotes the ego vehicle. The third column presents the representations of collaborative observation sequences, where the yellow areas denote the occluded objects that are not shown in the ego observations. The length of lines denotes the temporal sequence length.
}
\label{fig:qual}
\vspace{-12pt}
\end{figure}
\subsection{Results on Connected Autonomous Driving}
The CAD simulation contains a variety of challenges to perform connected autonomous driving, including complex interaction among street objects (e.g., vehicle yielding), strong occlusion in observations, limited communication bandwidth, missing objects in observation sequences, and highly dynamic situations. We run our approach on a Linux machine with an i7 16-core CPU, 16G memory, and RTX 3080 GPU. 
The average running speed is approximately 30 Hz. For each run, the spatiotemporal representation takes around 30 ms and the network forward process takes around 3 ms.

The quantitative results are presented in Table \ref{tab:QuanResults}. We can see that our approach performs the best on accident detection \textbf{AD} in three scenarios, which indicates the {\em best performance in guaranteeing safety in autonomous driving}. It is because of the capability of integrating multi-vehicle observations and temporal observations. In addition, our approach achieves {\bf over 100 times reduction} in the shared package size. This achievement substantially alleviates the communication bandwidth burden, rendering it highly suitable for real-world applications.
Furthermore, we can observe that \textbf{GAT} methods without considering temporal and collaborative observations perform worst among all comparisons. The \textbf{Raw Data Sharing} method performs better than \textbf{GAT} and \textbf{Collab-GAT}, which indicates the importance of integrating temporal cues and collaborative observations for decision-making. However, its package size for sharing is $6$ MB, which is far beyond the  standard $200$ KB and $720$ KB in DSRC and C-V2X. The \textbf{Compressed Feature Sharing} method performs better than the \textbf{Raw Data Sharing} method on both communication efficiency and driving performance. However, the compressed features still need the package size to be $510$ KB for a single observation. By efficiently encoding spatiotemporal cues, our baseline model outperforms all the other methods in most cases, especially on the improvements in driving safety indicated by the higher value of accident detection. Our full approach performs the best by further explicitly encoding distances and time gaps indicated by the edge attributes.

The qualitative results are presented in Figure \ref{fig:qual}. We can clearly observe that our approach can effectively detect accidents and take brake actions. It is because our proposed spatiotemporal graph representations can effectively preserve the spatiotemporal relationships of street objects in a communication-efficient way.
Furthermore, only using ego observation with strong occlusion may trigger accidents. By integrating collaborative observations, the occluded objects are correctly detected, thus the ego vehicle can make safe and effective decisions to avoid traffic accidents.

\begin{table}[t]
\vspace{8pt}
\centering
\tabcolsep=0.15cm
\caption{Ablation Analysis on Model Components:  Temporal coherency and data sharing 
}
\label{tab:ablation}
\begin{tabular}{|l|c|c|c|c|c|c|c|c|c|c|c|c|}
\hline
Method  & \multicolumn{2}{ c|}{Overtaking }  &  \multicolumn{2}{ c|}{Left Turn }  & \multicolumn{2}{ c|}{Street Crossing }\\
\cline{1-7}
		 & AD $\uparrow$ & EAR $\uparrow$ 
         & AD $\uparrow$ & EAR $\uparrow$ 
         & AD $\uparrow$ & EAR $\uparrow$ \\
		\hline
        NT-NS  
         &0.7588 & 0.6565 & 0.2272 & 0.6041 & 0.2960 & 0.7505\\
        \hline
	T-NS
         & 0.5873 & 0.7018  & 0.2271 & 0.6564 & 0.1762 & 0.7575\\
        \hline
        NT-S
         & 0.5983 &0. 7020 & 0.2561 & 0.6204  & 0.5018 & 0.7195\\

        \hline 
		  Full (Ours) 
         & \textbf{0.9265}  & \textbf{0.8336} & \textbf{0.6070} & \textbf{0.7670}  & \textbf{0.6451} & \textbf{0.7846}\\
        \hline
        
\end{tabular}\vspace{-8pt}
\end{table}
\subsection{Ablation Study}
We also conduct an ablation study to analyze the components of our approach. Specifically, we test our approach in the following scenarios:
\begin{itemize}
    \item No Temporal and No Sharing (\textbf{NT-NS}) that uses the same model as ours but only uses the ego vehicles' observation. In this case, the spatiotemporal graph is downgraded to a regular graph, thus using the traditional graph attention network to deal with it.
    \item No Temporal with Sharing (\textbf{NT-S}) that uses the same network as above to aggregate collaborator observations but without using observation sequences.
    \item Temporal but No Sharing (\textbf{T-NS}) that uses the same model as ours with temporal observations of ego vehicle and without using collaborators' observation sequences.
\end{itemize}

The ablation study results are presented in Table \ref{tab:ablation}. We can see that \textbf{NT-NS} performs the worst in the scenario of left turn and street crossing. By adding temporal and collaborative observations, \textbf{T-NS} and \textbf{NT-S} achieve better imitation performance (indicating by \textbf{EAR}) compared with \textbf{NT-NS},
further proving the importance of collaborative and temporal information for decision-making in connected autonomous driving. By considering both cues, our full approach achieves the best performance. 

\section{Conclusion}
We propose a novel approach for collaborative decision-making in connected autonomous driving to avoid traffic accidents. Our approach significantly reduces the sharing data size by representing a sequence of observations as a spatiotemporal graph. Then, design a novel spatiotemporal graph neural network based on heterogeneous graph learning to perform collaborative decision-making, which can analyze the spatial and temporal connection of objects in a unified way. Experimental results have shown that our approach outperforms the existing methods in communication efficiency, driving safety, and imitation performance.

Our approach has some limitations, offering possible future directions. First, we assume the transformation of collaborators' observations is based on an accurate GNSS sensor. Deploying our approach with noisy global coordination or in GPS-denied environments can be studied in the future. Second, the current spatiotemporal graphs only encode object topology information, we can further add more complex relationships into the graph representation given heterogeneous graph learning, such as adding lane information or geometric information of vehicles, to improve the performance.

\bibliographystyle{IEEEtran}
\bibliography{ref}
\end{document}